\documentclass[10pt,twocolumn,letterpaper]{article}

\usepackage[pagenumbers]{cvpr} % To force page numbers, e.g. for an arXiv version

\usepackage{graphicx}
\usepackage{amsmath}
\usepackage{amssymb}
\usepackage{booktabs}
\usepackage{verbatim}
\usepackage{multirow}

\usepackage[pagebackref,breaklinks,colorlinks]{hyperref}

\usepackage[capitalize]{cleveref}
\crefname{section}{Sec.}{Secs.}
\Crefname{section}{Section}{Sections}
\Crefname{table}{Table}{Tables}
\crefname{table}{Tab.}{Tabs.}

\def\cvprPaperID{5283} % *** Enter the CVPR Paper ID here
\def\confName{CVPR}
\def\confYear{2022}

\begin{document}

\title{Pseudo-Stereo for Monocular 3D Object Detection in Autonomous Driving}
% Pseudo-Stereo: Learning Virtual View for Monocular 3D Object Detection
%Look Ma, No Ground Truth Right Images: Pseudo-Stereo 3D Object Detection for Autonomous Driving

\author{Yi-Nan Chen$^1$\quad Hang Dai$^{2*}$\quad Yong Ding$^{1*}$\\
$^1$School of Micro-Nano Electronics, Zhejiang University\\
$^2$Mohamed bin Zayed University of Artificial Intelligence, Abu Dhabi, UAE
\\
{\tt\small $^*$Corresponding authors\{hang.dai@mbzuai.ac.ae, dingy@vlsi.zju.edu.cn\}.}
}

\maketitle

%%%%%%%%% ABSTRACT
\begin{abstract}
Pseudo-LiDAR 3D detectors have made remarkable progress in monocular 3D detection by enhancing the capability of perceiving depth with depth estimation networks, and using LiDAR-based 3D detection architectures. The advanced stereo 3D detectors can also accurately localize 3D objects. The gap in image-to-image generation for stereo views is much smaller than that in image-to-LiDAR generation. Motivated by this, we propose a Pseudo-Stereo 3D detection framework with three novel virtual view generation methods, including image-level generation, feature-level generation, and feature-clone, for detecting 3D objects from a single image. Our analysis of depth-aware learning shows that the depth loss is effective in only feature-level virtual view generation and the estimated depth map is effective in both image-level and feature-level in our framework. We propose a disparity-wise dynamic convolution with dynamic kernels sampled from the disparity feature map to filter the features adaptively from a single image for generating virtual image features, which eases the feature degradation caused by the depth estimation errors. Till submission (November 18, 2021), our Pseudo-Stereo 3D detection framework ranks 1$^{st}$ on car, pedestrian, and cyclist among the monocular 3D detectors with publications on the KITTI-3D benchmark. The code is released at \url{https://github.com/revisitq/Pseudo-Stereo-3D}.%Our code will be released.

\end{abstract}

%%%%%%%%% BODY TEXT

% Pesudo Stereo VS Pesudo LiDAR, Pesudo Stereo relies on the development of image-to-image generation which is better explored than image-to-point generation.
% Monocular learn from stereo in deep feature space
%伪双目用于单目检测，
%双目有很好，用到单目关键是如何生成image features for 
% image-to-image, feature-to-feature, feature clone
\vspace{-4mm}
\section{Introduction}
\label{sec:intro}
% 1. Shallow learning (2D backbone + 3D detection) 
% 2. Advanced learning--compose 3D feature learning: (1) from left image and virtual right image; (2) from left featureand the right featuregenerated by depth map feature offsets + left featureto generate 3D feature volume; (3) directly from single left image to generate 3D feature volume with depth guidance loss; 

% Types of monocular 3D detector:
% input 2D left image                                                    + 2D backbone   + 3D detection head
% input 2D left image (and optional predicted depth map from left image) +2D-3D backbone + 3D detection head
% input Pseudo LiDAR from 2D left image                                  + 3D backbone   + 3D detection head

%总的单目结构就是这三类，中间这一类是我们论文要探讨的，predicted depth是可供选择的输入，后面接的是2D-3D backbone + detection head，这一类有三种子类型：（1）纯 left； （2） left + virtual right image generated from predicted depth map; (3) left feature+ generated right feature from predicted depth;
%introduction会从总的单目结构讲起，引入这种类型，讨论它相比其他两大类的优势，然后介绍探索三个子类型
% contribution: （1）在2D-3D backbone 作为非常promising的单目检测方向上，核心关键是怎么获得3D feature volume, 探索三种3D feature learning的子类型，得出最优的结构，并分析原因； （2）predicted depth 作为输入的影响； （3）depth loss的影响

%3D feature learning 
Detecting the 3D objects from monocular image enables the machine to perceive and understand the 3D real world, which has a wide applications including virtual reality, robotics and autonomous driving. Monocular 3D detection is a challenging task because of the lack of accurate 3D information in a single image. However, the huge potential in such a cheap and easy-to-deploy solution to 3D detection attracts more and more researchers. 
Remarkable progress has been made in Pseudo-LiDAR detectors \cite{Ma_2020_ECCV,you2020pseudo,Liang_corr,simonelli2021we,chu2021neighbor} that use a pre-trained depth estimation network to generate Pseudo-LiDAR representations, $e.g.$ pseudo point clouds and pseudo voxels, and then feed them to LiDAR-based 3D detectors. It shows that enhancing the capability of perceiving depth can improve monocular 3D detection performance. However, there is a huge performance gap between Pseudo-LiDAR and LiDAR-based detectors because of the errors in image-to-LiDAR generation \cite{dd3d}.  

Apart from LiDAR-based detectors, the stereo 3D detectors \cite{guo2021liga,chen2020dsgn} can also accurately localize 3D objects. Also, the gap in image-to-image generation for stereo views is much smaller than that in image-to-LiDAR generation, which is a cross-modality conversion. Instead of Pseudo-LiDAR, we propose a novel Pseudo-Stereo 3D detection framework for monocular 3D detection. Our Pseudo-Stereo 3D detection framework generates a virtual view from a single input image to compose Pseudo-Stereo views with the generated virtual view and the input view. Then, we feed the Pseudo-Stereo views to stereo 3D detectors for detecting 3D objects from the single input image. We use one of the most advanced stereo 3D detectors, LIGA-Stereo \cite{guo2021liga}, as the base detection architecture. Thus, the virtual view generation is the key to our Pseudo-Stereo 3D detection framework.

%From the progress in monocular depth estimation \cite{poggi2020uncertainty,johnston2020self,ye2021unsupervised}, the 3D feature volume construction is the key to 3D perception from images \cite{zhao2020monocular}.

\begin{figure}[t!]
  \centering
 \includegraphics[width=0.48\textwidth]{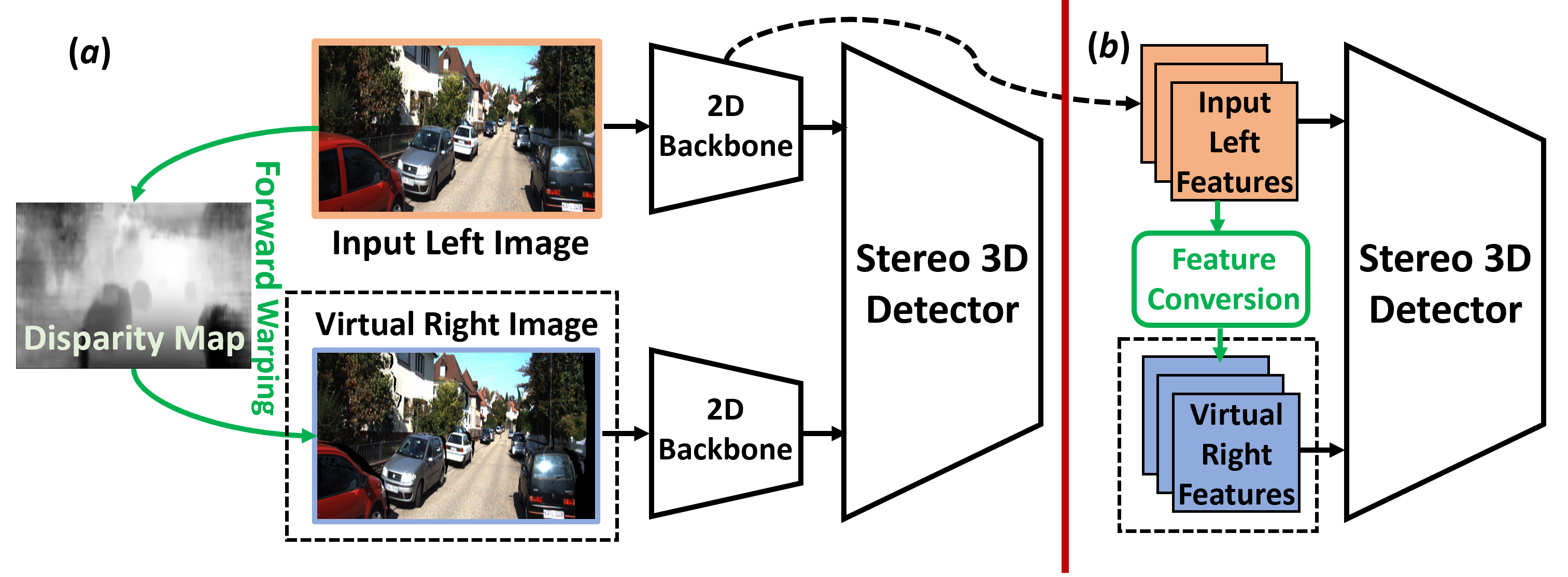}
\vspace{-5mm}
  \caption{Overview of our Pseudo-Stereo 3D detection framework with novel virtual view generation methods: (a) Image-level to use the generated disparity map for forward warping the input left image into a virtual right image, (b) Feature-level to convert the left features into virtual right features. A feature conversion baseline is to clone the left features as the special case in stereo views.
} 
 \label{fig:virtual}
\vspace{-5mm}
\end{figure}

We take KITTI-3D as an example only to explain how to generate a virtual view. Note that the virtual view does not require the ground-truth actual view in the dataset for training. In KITTI-3D, the monocular 3D detection is performed on the left image from the stereo views. Our aim is to construct Pseudo-Stereo views by generating the virtual right view from the input left view in either image-level or feature-level for monocular 3D detection. As shown in Figure~\ref{fig:virtual}, we propose two types of virtual view generation: (a) image-level to generate the virtual right image from the input left image and (b) feature-level to convert the left features into virtual right features. In image-level, we convert the estimated depth map from the input left image into disparities and use them to forward warp the input left image into a virtual right image to compose the Pseudo-Stereo views with the input left view. In feature-level, we propose a disparity-wise dynamic convolution with dynamic kernels sampled from disparity feature map to filter the left features adaptively for generating virtual right features, which eases the feature degradation caused by the depth estimation errors.
Also, a simple feature conversion is to clone the left features as the virtual right features, which is the special case of stereo views that the virtual right view is the same as the left view. We summarize our \textbf{contributions}:
\begin{itemize}
\vspace{-3mm}
    \item We propose a Pseudo-Stereo 3D detection framework with three novel virtual view generation methods, including image-level generation, feature-level generation and feature-clone, for detecting 3D objects from a single image, achieving significant improvements in monocular 3D detection. The proposed framework with feature-level virtual view generation ranks 1$^{st}$ among the monocular 3D detectors with publications across three object classes on KITTI-3D benchmark.
    \vspace{-3mm}
    \item In our framework, we analyze two major effects of learning depth-aware feature representations, including the estimated depth map and the depth loss as the depth guidance. It is very interesting to find that the depth loss is effective in feature-level virtual view generation only and the estimated depth map is effective in both image-level and feature-level for depth-aware feature learning.    
    \vspace{-3mm}
    \item In our feature-level virtual view generation method, we propose a disparity-wise dynamic convolution with dynamic kernels from disparity feature map to adaptively filter the features from a single image for generating virtual image features, which avoids the feature degradation caused by the depth estimation errors.
\end{itemize}

% for monocular 3D detection pseudo 3D data representation learning first converts the estimated depth map into intermediate pseudo 3D data which is then fed to LiDAR-based networks \cite{yan2018second}. Thus, the depth estimation errors account for the huge performance gap between pseudo 3D data and LiDAR-based methods \cite{dd3d}. To resolve the issue that 2D features have limited capability of 3D perception from two dimensional spatial expressions, some methods use large-scale monocular depth pre-training \cite{dd3d} or the estimated depth guidance \cite{ding2019learning} to learn features that are sensitive to depth information. From the progress in monocular depth estimation \cite{poggi2020uncertainty,johnston2020self,ye2021unsupervised}, constructing 3D feature volume from single image is crucial to 3D perception from images \cite{zhao2020monocular}. In this paper, we propose three novel Pseudo-Stereo methods to construct 3D feature volume from single image and study the effects of the major components in the proposed Pseudo-Stereo methods for monocular 3D detection.

\section{Related Works}
The architectures for monocular 3D object detection can be mainly categorized into two groups:
Pseudo-LiDAR based methods \cite{Liang_corr,simonelli2021we,chu2021neighbor} that use pre-trained depth networks to
generate pseudo LiDAR representations, $e.g.$ pseudo point clouds and pseudo voxels, and then feed them to LiDAR-based 3D detectors, and the rest monocular 3D detection methods that use 2D feature learning from a single image with optional 3D cues matching, concatenating or guiding for 3D perception \cite{Lu_2021_ICCV,liu2021autoshape,MonoFlex,Zou_2021_ICCV,reading2021categorical,rukhovich2021imvoxelnet,murez2020atlas}. %We denote the above three types feature learning architectures as \textbf{Type} \textbf{I}, \textbf{Type} \textbf{II} and \textbf{Type} \textbf{III} in the following literature.

\textbf{Monocular 3D Detection.} There are a few monocular 3D detectors use 2D feature learning in 2D backbone with optional 3D cues concatenated or matched to 2D features for 3D perception.
Chabot $\etal$ \cite{chabot2017deep} estimate the similarity between the detected vehicle and a pre-defined 3D vehicle shape template used as 3D cues in 2D backbone. They solve the 3D location and 3D rotation angle of the detected vehicle in a standard 2D/3D matching algorithm \cite{lepetit2009epnp}. Barabanau $\etal$ \cite{barabanau2019monocular} use 2D features to predict the rotation and key points of a car in a 2D backbone. Then, they use a geometric reasoning between the key points and the corresponding points in CAD models to get the depth and 3D locations of the car. But it is difficult to get CAD models of all object classes.
GrooMeD-NMS \cite{kumar2021groomed} extracts 2D features for monocular 3D detection, with a differentiable NMS selecting the best 3D box candidate. GS3D \cite{li2019gs3d} uses a specifically designed 2D backbone to extract the surface features for tackling the representation ambiguity between 2D bounding box and 3D bounding box. MonoEF \cite{zhou2021monocular} employs a 2D backbone with a camera extrinsic parameter aware module to deconstruct camera extrinsic parameters from 3D detection parameters. M3D-RPN \cite{brazil2019m3d} extracts 2D image feature to predict both 2D and 3D bounding boxes directly by minimizing the distance error between the 2D projection of the predicted 3D bounding box and the predicted 2D bounding box. Following M3D-RPN \cite{brazil2019m3d}, many works \cite{luo2021m3dssd,park2021pseudo,peng2021lidar} enhance the 2D feature learning with the 2D-3D detection head for monocular 3D detection. 

Some methods aggregate the 2D image feature and the depth features extracted from the depth map to get 2D depth-aware features \cite{ding2019learning,park2021pseudo}. D$4$LCN \cite{ding2019learning} employs a depth-guided convolution with the weights and the receptive fields learning from the estimated depth for depth-aware feature extraction. DDMP-3D \cite{wang2021depth} uses a depth-conditioned propagation based on graph to learn 2D depth-aware features for monocular 3D detection. DD3D \cite{park2021pseudo} adds a depth prediction head to the 3D detection head and uses a depth loss to learn 2D features that are sensitive to depth information for monocular 3D detection. DD3D\cite{park2021pseudo} also pre-trains the depth prediction head on a large-scale dataset and fine-tunes the overall network on monocular 3D detection task.
Other methods extract 2D features and construct 3D feature volume from the transformation of 2D features to improve 3D perceiving capacity. CaDDN \cite{reading2021categorical} uses the estimated depth distributions to construct a frustum feature grid. Then, the frustum feature is converted into a voxel grid using known camera calibration parameters to construct 3D voxel feature volumes. ImVoxelNet \cite{rukhovich2021imvoxelnet} uses 2D backbone to extract 2D image feature and projects the 2D features into 3D feature volumes following \cite{murez2020atlas}. Then, the 3D feature volumes go through 3D backbone to enhance the 3D features for monocular 3D detection.

\textbf{Pseudo-LiDAR.} Pseudo-LiDAR architecture converts the estimated depth map from a single image into Pseudo 3D data representations \cite{vianney2019refinedmpl,cai2020monocular} which are then fed to 3D backbone to learn point-wise, voxel-wise or bird's eye view (BEV) features for monocular 3D detection. RefinedMPL \cite{vianney2019refinedmpl} uses PointRCNN \cite{shi2019pointrcnn} for point-wise feature learning in a supervised or an unsupervised scheme from pseudo point clouds prior. AM3D \cite{ma2019accurate} uses a PointNet \cite{qi2017pointnet} backbone for point-wise feature extraction from pseudo point clouds, and employs a multi-modal fusion block to enhance the point-wise feature learning. MonoFENet\cite{bao2019monofenet} enhances the 3D features from the estimated disparity for monocular 3D detection. Decoupled-3D \cite{cai2020monocular} recovers the missing depth of the object using the coarse depth from 3D object height prior with the BEV features that are converted from the estimated depth map. However, the performance and the generalization capability of these methods rely on the accuracy of image-to-LiDAR generation, which has a huge gap between the two data modalities. 

%replace the stereo image extraction block with our Pseudo-Stereo image feature generation block
\section{Preliminaries of Stereo 3D Detector}\label{sec:liga}
Volume-based stereo 3D detectors aim to generate 3D anchor space from stereo image \cite{qin2019triangulation} and localize 3D objects from 3D feature volume \cite{chen2020dsgn,wang2021plume,guo2021liga}. DSGN \cite{chen2020dsgn} follows the widely used 3D cost volume construction in stereo matching \cite{xu2020aanet,tankovich2021hitnet,gu2020cascade} with 3D geometric information encoding. The depth loss in stereo matching branch helps learn depth-aware features for the detection branch, improving the detection accuracy. Wang $\etal$ \cite{wang2021plume} use a direct construction of 3D cost volume to reduce the computational cost. Based on DSGN \cite{chen2020dsgn}, LIGA-Stereo \cite{guo2021liga} achieves significant improvements against other methods \cite{chen2020dsgn,PL++,bewley2020range} in stereo 3D detection. Thus, we use LIGA-Stereo \cite{guo2021liga} as our base stereo 3D detection architecture and feed the Pseudo-Stereo views to LIGA-Stereo. We focus on how to generate the virtual right view from the input left view and learn Pseudo-Stereo features that are sensitive to depth information. Thus, we introduce the stereo image feature extraction and the 3D feature volume construction in LIGA-Stereo \cite{guo2021liga}. 

\textbf{Stereo Image Feature Extraction.}
Given an image pair $(I_L, I_R)$ from stereo views, the LIGA-Stereo \cite{guo2021liga} first extracts the left features and the right features via a ResNet-34 \cite{he2016deep} with shared weights as the 2D image backbone.
The strides of the output feature map in the five blocks are 2, 2, 4, 4 and 4, respectively. The channels of the output feature map in the five blocks are 64, 64, 128, 128 and 128. Then, we denote the left and the right features as $F'_L$ and $F'_R$ that are the input to the spatial pyramid pooling (SPP) module \cite{chang2018pyramid} with shared weights for getting the final left features $F_L$ and right features $F_R$. The strides of the final left features $F_L$ and right features $F_R$ are 1, and the channels of the final left features $F_L$ and the right features $F_R$ are 32.
%of $conv\_2\sim conv\_5$ 
%We denote the five convolution blocks of the modified ResNet-34 \cite{he2016deep} as $conv\_1$, $conv\_2$, ..., $conv\_5$.

\textbf{The 3D Feature Volume Construction. }
With the left features $F_L$ and the right features $F_R$, the stereo volume $V_{st}$ is built by concatenating the left features $F_L$ with the re-projected right features $F_{R->L}$ at every candidate depth level. Thus, the stereo volume construction can be formulated with camera parameters as:
\begin{eqnarray}
&V_{st}(u,v,w) = concat[F_L(u,v), F_{R->L}(u, v)]\\
&F_{R->L}(u, v) = F_R(u - \frac{f \cdot b}{d(w)\cdot S}, v)\\
&d(w)=w \cdot v_d + z_{min}
\end{eqnarray}
where $(u,v)$ are the pixel coordinates, $w \in [0, 1, ...]$ indicates the depth index, $S$ is the stride of the feature map, $v_d$ is the depth interval, $z_{min}$ indicates the minimal depth value, $f$ is the camera focal length, and $b$ represents the baseline of the stereo camera pair. 
After the stereo volume $V_{st}$ is filtered by a stereo network 3D Hourglass \cite{guo2021liga}, we get a re-sampled stereo volume $V'_{st}$ and a depth distribution volume $P_{st}$. The $P_{st}$ describes the depth probability distribution of pixels for all the candidate depth levels described in $d(w)$. A \textbf{depth loss} is computed between the depth map regressed from the re-sampled stereo volume $V'_{st}$ and the ground-truth depth map to guide the depth-aware learning of $V'_{st}$. With the camera calibration parameters, we can transform the volume in stereo space $V'_{st}$ to the volume in 3D space $V_{3d}$ by concatenating the semantic features from left image penalized by the depth probability $P_{st}$ and the re-sampled stereo volume $V'_{st}$. 
Following SECOND \cite{yan2018second}, the 3D feature volume $V_{3d}$ is collapsed to a bird's eye view (BEV) feature map $F_{BEV}$ by merging the dimension of height and channel. Finally, a 2D aggregation network is used to get a refined BEV feature map $F'_{BEV}$ that is used for the regression of 3D detection parameters.

\begin{figure*}[t!]
  \centering
 \includegraphics[width=1\textwidth]{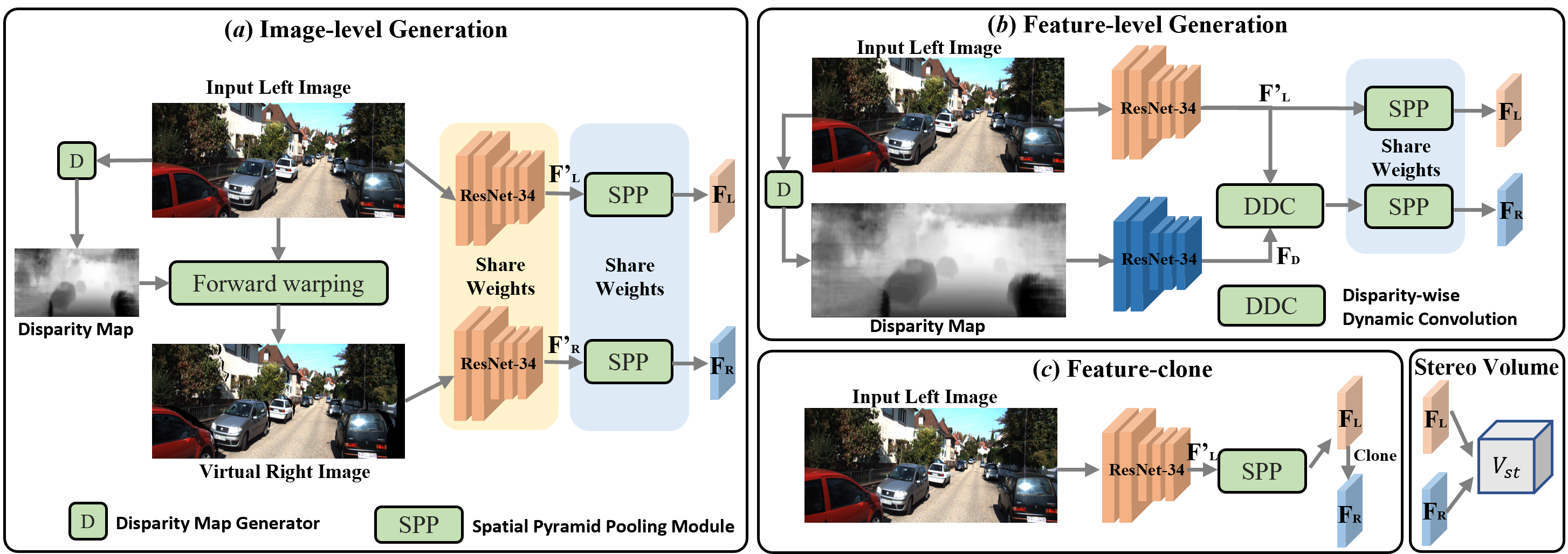}
\vspace{-6mm}
  \caption{ Overview of our virtual view generation methods: (a) Image-level that uses the generated disparity map for forward warping the input left image into a virtual right image, (b) Feature-level that converts the left features into virtual right features via the proposed disparity-wise dynamic convolution (DDC), (c) Feature-clone that simply duplicates the left features as the virtual right features.
} 
 \label{fig:overview}
\vspace{-3mm}
\end{figure*} 

\section{Method}
As shown in Figure~\ref{fig:overview}, we propose three novel methods to generate the virtual right view from the input left view and construct the Pseudo-Stereo views in (a) image-level in Section~\ref{sec:img2img}, (b) feature-level in Section~\ref{sec:img2feat}, and (c) feature-clone as the baseline of feature-level generation in Section~\ref{sec:img}. We describe the loss function in Section~\ref{sec:loss}. In Section~\ref{sec:depth_loss}, we analyze the depth-aware feature learning in the proposed Pseudo-Stereo 3D detection framework.

\subsection{Pseudo-Stereo 3D Detection Framework}
We use LIGA-Stereo \cite{guo2021liga} as our base stereo 3D detection architecture and replace the stereo image feature extraction block with our Pseudo-Stereo image feature extraction block. As shown in Figure~\ref{fig:overview}, we propose three virtual right view generation methods and extract Pseudo-Stereo image features from the input left view and the generated virtual right view. Then, we feed the Pseudo-Stereo image features to LIGA-Stereo for detecting 3D objects from the input left image only. 
%We focus on how to generate the image features for the virtual right view from the input left image and improve the capability of perceiving depth in our framework.

\subsection{Image-level Generation}\label{sec:img2img}
\begin{figure}[t!]
    \centering
    \begin{subfigure}{\columnwidth}
    \centering
        \includegraphics[width=0.95\textwidth]{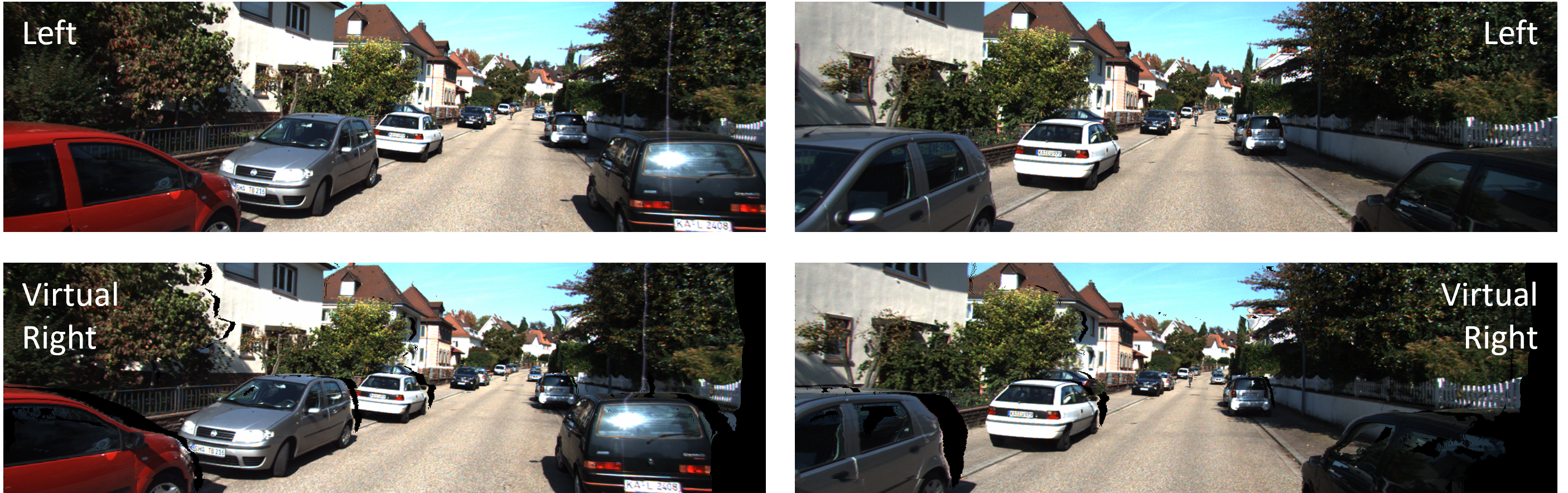}
    \end{subfigure}
    \vspace{-3mm}
    \caption{\small Left image (top) and the generated virtual right image (bottom) using our image-level virtual view generation method.}
    \label{fig:left2right}
    \vspace{-4mm}
\end{figure}

In image-level, we generate a virtual right image $\hat{I}_R$ from the input left image ${I}_L$ using the estimated disparity map as shown in Figure~\ref{fig:overview}(a). Then, we extract Pseudo-Stereo image features from the Pseudo-Stereo pair $({I}_L, \hat{I}_R)$. 
%We use a pre-trained DORN \cite{fu2018deep} to estimate depth map $Z$ from the input left image.
With a pair of pixel correspondences $x_l$ and $x_r$ in the left image $I_L$ and the right image $I_R$,  
the disparity $d$ between the pair of pixel correspondences can be computed as:
\begin{equation}\label{eqn:d}
d = x_l - x_r
\end{equation}
Given the depth value $z$ for the pixel $x_l$ in the input left image and the camera calibration parameters,
the relationship between the disparity $d$ and its corresponding depth value $z$ can be formulated as:
\begin{equation}\label{eqn:z2d}
d = \frac{f \cdot b}{z}
\end{equation}
where $f$ and $b$ are the camera focal length and the baseline of the stereo camera pair, respectively.

To get the virtual right image, we first use a pre-trained DORN \cite{fu2018deep} to estimate the depth map $Z$ from the input left image $I_L$. Then, we convert the depth map $Z$ to a disparity map $D$ according to Eqn.~\ref{eqn:z2d} with camera parameters. Following Eqn.~\ref{eqn:d}, we use the disparity map to forward warp the left image $I_L$ \cite{schwarz2007non} into the virtual right image $\hat{I}_R$ as shown in Figure~\ref{fig:left2right}. To address ‘blurry’ edges, occlusions and collisions, we sharpen the disparity map by identifying the flying pixels and applying a Sobel edge filter response of greater than 3 to the disparity map on those flying pixels \cite{hu2019revisiting,watson2020learning}. In image-level generation, we embed the estimated depth information extracted from the left image into the virtual right image for Pseudo-Stereo 3D detection. Then, we use a ResNet-34 \cite{he2016deep} with shared weights followed by a spatial pyramid pooling (SPP) module with shared-weights to extract the left features $F_L$ and the virtual right features $\hat{F}_R$ from the Pseudo-Stereo pair $({I}_L, \hat{I}_R)$. We can use the Pseudo-Stereo image features to construct the stereo volume $V_{st}$ as detailed in Section~\ref{sec:liga}.   

\subsection{Feature-level Generation}\label{sec:img2feat}
Generating the virtual right image is a time-consuming process because of the forward warping \cite{hu2019revisiting,watson2020learning}. To overcome this, we propose a differentiable feature-level method for generating the virtual right features from the left features and the disparity features as shown in Figure~\ref{fig:overview}(b). We convert the estimated depth map into a disparity map and use two ResNet-34 \cite{he2016deep} to extract the left features $F'_L$ from the left input image and the disparity features $F_D$ from the disparity map. The two ResNet-34 are not with shared weights. %Specifically, we use the left features and the disparity features from $conv\_3\sim conv\_5$ to generate virtual right features.

\begin{figure}[t!]
    \centering
    \begin{subfigure}{\columnwidth}
        \includegraphics[width=\textwidth]{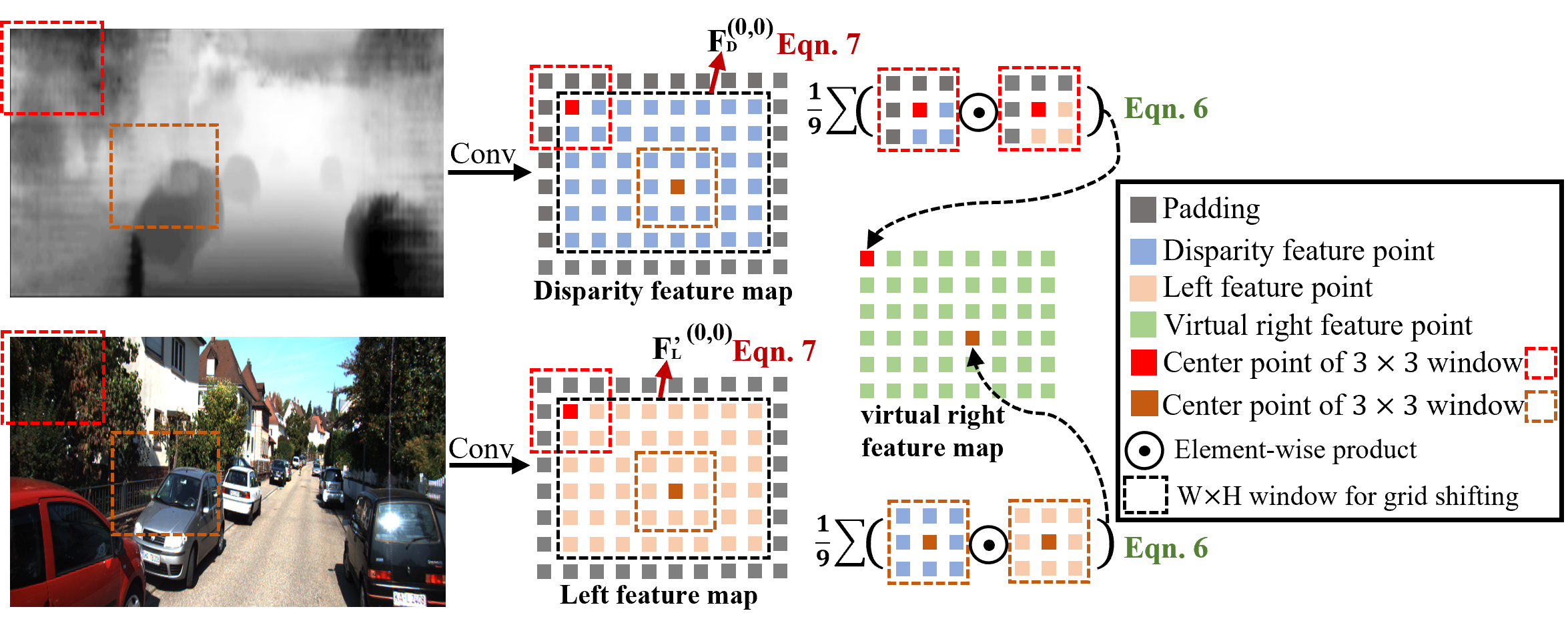}
    \end{subfigure}
    \vspace{-3mm}
    \caption{\small An illustration of disparity-wise dynamic convolution.}
    \label{fig:conv}
    \vspace{-3mm}
\end{figure}

Instead of computing the offsets to compensate the left view as the virtual right view, we propose a disparity-wise dynamic convolution (DDC) to filter the left feature map $F'_L\in\mathbb{R}^{W\times H \times C}$ adaptively by the dynamic kernels from the disparity feature map $F_D\in\mathbb{R}^{W\times H \times C}$ for generating the virtual right feature map $\hat{F'}_R\in\mathbb{R}^{W\times H \times C}$, where $W$, $H$ and $C$ are the width, height and channel of the feature map, respectively. As shown in Figure~\ref{fig:conv}, the adaptive filtering process use a $3 \times 3$ sliding window to cover all the feature points in $F_D$ and $F'_L$:
\begin{eqnarray}
\vspace{-2mm}
&\hat{F}_R(i, j) =\frac{1}{3 \times 3} \sum\limits_{u'}\sum\limits_{v'}F_D(u', v') \cdot F'_L(u',v')
\vspace{-2mm}
\end{eqnarray}
where $u' \in \{i-1, i, i+1\}$ and $v' \in \{j-1, j, j+1\}$ are the coordinates of the feature points in the sliding window. We need to apply $W*H$ times sliding window to cover the whole feature map, which is not efficient. Instead, we use a grid shifting operation that can cover the whole feature map with 9 times shifting. After padding, we shift a $W \times H$ window following the direction and the step size represented in a $3\times3$ grid $\{(g_i, g_j)\}$, $g \in\{-1,0,1\}$ on $F_D$ and $F'_L$ for getting the kernel $F_D^{(g_i, g_j)} \in \mathbb{R}^{W\times H \times C}$ and the feature map  $F'^{(g_i, g_j)}_L \in \mathbb{R}^{W\times H \times C}$, respectively. When $g_i = 0$ and $g_j = 0$, the $W \times H$ window covers the original feature map that is without padding as shown in black dot box of Figure~\ref{fig:conv}. Thus, we can get the virtual right features $\hat{F}'_R$ by filtering $F'^{(g_i, g_j)}_L$ adaptively by the kernels $F_D^{(g_i, g_j)}$:
\begin{eqnarray}
\vspace{-4mm}
%&\hat{F}'_R = F'_L \odot \frac{1}{3\times 3}\sum\limits_{g_i, g_j}F_D^{(g_i, g_j)} 
&\hat{F}'_R =  \frac{1}{3\times 3}\sum\limits_{g_i, g_j}F'^{(g_i, g_j)}_L \odot F_D^{(g_i, g_j)} 
\vspace{-4mm}
\end{eqnarray}
where the grid shifting operation is applied nine times to cover the whole feature map. For more details, please refer to \textcolor{blue}{supplementary materials}. We feed $F'_L$ and $\hat{F'}_R$ to SPP module with shared weights using strides of 4 for getting the final left features $F_L$ and virtual right features $\hat{F}_R$.% which are used for monocular 3D detection.   
%we use a $3 \times 3$ dynamic kernel sampled from the disparity feature map to filter the corresponding position of a $3 \times 3$ left feature block.
%After padding, we use 3*3 pieces of $W \times H$ kernel from the disparity feature map $F_D$ generated by shifting with the guidance of the grid $\{(g_i, g_j)\}$, $g \in\{-1,0,1\}$ as $F_D^{(g_i, g_j)} \in \mathbb{R}^{W\times H}$. The grid indicates the direction and the step size of the shifting operation. We apply the kernel $F_D^{(g_i, g_j)}$ to the corresponding position on the left features denoted as $F'^{(g_i, g_j)}_L \in \mathbb{R}^{W\times H}$.

%But the weights of the kernel is dynamic generated from the disparity feature. We use a dynamic local filter that proposed by D4LCN\cite{ding2019learning} to implement this operation.
Compared with the image-level generation, the feature-level generation is faster without forward warping and more adaptive using the proposed disparity-wise dynamic convolution.
Also, by embedding the estimated depth information into high dimensional feature space and using the embedded depth information to filter the left features, it mitigates the degradation of depth-aware representations and the depth-aware representations can strengthen the embedded depth information with extra depth guidance, achieving significant improvements in monocular 3D detection. 

\subsection{Image Feature Clone}\label{sec:img}
\vspace{-1mm}
We clone the left features as the virtual right features as shown in Figure~\ref{fig:overview}(c). This can be seen as the special case of Pseudo-Stereo views that the virtual right view is the same as the left view. Also, feature cloning is used as the baseline of feature-level generation. With different Pseudo-Stereo views, the proposed framework can improve the representations of 3D feature volume with the paired pixel-correspondence constraints or the feature-correspondence constraints converted from the estimated depth map. However, cloning feature does not need a pre-trained depth estimation network in our Pseudo-Stereo 3D detection framework, leading to better generalization capability.   

%Compared to strategy I and strategy II, strategy III is a more end-to-end manner because it doesn't need an additional depth estimator to generate the depth map.

%. our work is mainly study how to generate the right feature from left image and solve the monocular 3D detection task by a stereo-based framework, so we only modify the 2D feature extraction part.
\subsection{Loss Function}
\label{sec:loss}
Since we use LIGA-Stereo \cite{guo2021liga} as our base stereo 3D detection architecture and replace the original stereo image feature extraction block with the proposed three variants of Pseudo-Stereo image feature generation block as shown in Figure~\ref{fig:overview}, we employ the same loss function as LIGA-Stereo \cite{guo2021liga}, including the detection loss $L_{d}$ for the regression of all detection parameters, the depth loss $L_{depth}$ as the additional depth guidance for the re-sampled stereo volume $V'_{st}$ and the knowledge distillation loss $L_{kd}$ to transfer the structural knowledge from a LiDAR-based detector as described in LIGA-Stereo \cite{guo2021liga}. The overall loss can be formulated as:
\begin{eqnarray}
&L = \lambda _{d}L_{d}+ \lambda _{dep}L_{depth}+\lambda _{kd}L_{kd}
\end{eqnarray}
where $\lambda _{d}, \lambda _{dep}$, and $\lambda _{kd}$ are the regularization weights for the detection loss, the depth loss, and the knowledge distillation loss, respectively. The knowledge distillation adopted in LIGA-Stereo is well studied in \cite{guo2021liga}. Please refer to LIGA-Stereo \cite{guo2021liga} for more details. We focus on how to generate the virtual right view from the input left view and improve the capability of perceiving depth in features.  
%Depth-aware Feature Learning (1) Estimated depth as input (2) depth loss 
% mitigate gap 
\subsection{Learning Depth-aware Features}\label{sec:depth_loss}
%Influence Of Depth Loss In Three Strategies
The depth-aware feature learning in our framework lies in two aspects: the estimated depth map and the depth loss. we convert the estimated depth map as the disparity map and use it in either image space or feature space. By comparing the performance of the two methods, we can study the effect of the estimated depth map used for depth-aware feature learning in both image-level and feature-level. The depth loss $L_{depth}$ is used as the additional depth guidance for the re-sampled stereo volume $V'_{st}$ to improve the depth awareness in features, improving monocular 3D detection performance. Although both the estimated depth map and the depth loss can improve the depth awareness in feature learning, the interaction between the two factors is not well studied for monocular 3D detection before this work.  

%The depth loss in LIGA-Stereo\cite{guo2021liga} can help the 3D volume learn depth better. It's works well because the left-right feature pair that used to build the 3D volume is derived from the true left-right image pair. The reliable depth information can be inferred from the pixel-correspondence constraint between the left image and the right image. So when a supervision is applied on the depth learning, the network can learn depth better.
For the image-level generation in Section~\ref{sec:img2img}, we use a Pseudo-Stereo image pair to extract the Stereo image features, where the virtual right image is generated from the left image with the help of the estimated depth map. The monocular depth estimation is an ill-posed problem, which makes it difficult to get high-quality depth maps for virtual right image generation.
Thus, the pixel-correspondence constraints of the generated Pseudo-Stereo pairs may have large offsets against the ground truth because of the depth estimation errors.
Learning with the virtual right image warped from the inaccurate pixel-correspondences causes feature degradation. Since there is a huge gap between the ground-truth depth and the degraded feature, forcing the network to fit the ground-truth depth map using the depth loss with the degraded feature impairs the overall performance. For feature-level generation in Section~\ref{sec:img2feat}, the virtual right features are generated from the left features and the disparity features. Unlike image-level generation, where the forward-warping is a non-learning process from image to image, the feature-level generation is an adaptive learning process with disparity-wise dynamic convolutions from feature to feature. Also, the estimated depth information is embedded into high dimensional feature space and the embedded depth information is used to filter the left features in the feature-level generation. This eases the degradation of depth-aware representations, mitigating the gap between the ground-truth depth and the feature. Thus, the feature representations can evolve and refine the depth information with extra depth guidance, for example, a depth loss. For feature-clone in Section~\ref{sec:img}, we duplicate the left features as the Pseudo-Stereo image features without the estimated depth map. The depth loss alone can improve the depth awareness of features, improving the detection performance.

\section{Experiments}

\subsection{Dataset and Evaluation Metric}

\noindent\textbf{Dataset.} KITTI 3D object detection benchmark \cite{KITTI} is the most widely used benchmark for 3D object detection. It comprises 7481 training images and 7518 test images, along with the corresponding point clouds captured around a mid-size city from rural areas and highways. 
KITTI-3D provides 3D bounding box annotations for 3 classes, \emph{Car}, \emph{Cyclist} and \emph{Pedestrian}.
Commonly, the training set is divided into training split with 3712 samples and validation split with 3769 samples following that in \cite{ding2019learning}, which we denote as KITTI\,$train$ and KITTI\,$val$, respectively. All models in ablation studies are trained on the KITTI\,$train$ and evaluated on KITTI\,$val$. For the submission of our methods, the models is trained on the 7481 training samples.
Each object sample is assigned to a difficulty level, Easy, Moderate or Hard according to the object's bounding box height, occlusion level and truncation.

\noindent\textbf{Evaluation Metric.} We use two evaluation metrics in KITTI-3D, $i.e.$, the IoU of 3D bounding boxes or BEV 2D bounding boxes with average precision (AP) metric, which are denoted as $AP_{3D}$ and $ AP_{BEV}$, respectively. Following the monocular 3D detection methods\cite{ding2019learning, barabanau2019monocular, MonoFlex}, we conduct the ablation study on \emph{Car}. KITTI-3D uses the $AP|_{R40}$ with 40 recall points instead of $AP|_{R11}$ with 11 recall points from October 8, 2019. We report all the results in $AP|_{R40}$.
\subsection{Experiment Settings}
\noindent\textbf{Input Setting.} We use the pre-trained model of DORN \cite{DORN} to estimate the depth map. Then, we transform the depth maps into disparity maps with the camera calibration parameters. The virtual right images in image-level generation are generated before training to reduce the training time. For feature-level generation, the disparity map is normalized by $\mu$= 33.20, $\sigma$=15.91. The $\mu$ and $\sigma$ indicate the mean and variance of disparity map calculated from the training set.%before input it into the CNN backbone.

\noindent\textbf{Training Details.} The network is trained with an AdamW \cite{loshchilov2017decoupled} optimizer, with $\beta _1$=0.9, $\beta _2$=0.999. We train the network with 4 NVIDIA RTX 3090 GPUs. The batch size is set to 4. For the regularization weights of the training loss, $\lambda_{d}$=1.0, $\lambda_{kd}$=1.0. The regularization weight $\lambda _{dep}$ for depth loss $L_{depth}$ is set to 0 or 1, representing whether the depth loss is used or not. We use a single model to detect objects in different classes (\emph{Car}, \emph{Cyclist} and \emph{Pedestrian}) together. Other hyper-parameters are set as the same as LIGA-Stereo \cite{guo2021liga}. 

\begin{table}[htbp]
\vspace{-1mm}
    \small
    \centering
    \resizebox{\columnwidth}{!}{
    \begin{tabular}{c|c|c|ccc}
    \toprule
    \multirow{2}*{Exp.} &\multirow{2}*{Methods} &\multirow{2}*{$L_{depth}$} & \multicolumn{3}{c}{$AP_{3D}/AP_{BEV}$}  \\
    {} &{} &{} &Easy   &Moderate &Hard \\
    \midrule
    1&Image-level& {\checkmark} & {31.43 / 41.82} & {21.53 / 29.00} & {18.47 / 25.21} \\
    2&Image-level& {} &  {\textbf{31.81} / \textbf{42.87}} & {\textbf{22.36} / \textbf{30.16}} & {\textbf{19.33} / \textbf{26.38}} \\
    \midrule
    3&Feature-level& {\checkmark} &  {\textbf{35.18} / \textbf{45.50}} & {\textbf{24.15} / \textbf{32.03}} &{\textbf{20.35} / \textbf{27.57}}\\
    4&Feature-level& {} &  {22.04 / 31.10} & {16.18 / 22.55} &{14.31 / 20.56}\\
    \midrule
    5&Feature-clone& {\checkmark} &  {\textbf{28.46} / \textbf{37.66}} & {\textbf{19.15} / \textbf{25.78}} & {\textbf{16.56} / \textbf{22.47}} \\
    6&Feature-clone& {} &  {24.33 / 32.99} & {17.09 / 23.77} & {14.61 / 20.81} \\ 
    \bottomrule
    \end{tabular} }
    \vspace{-3mm}
  \caption{Ablation studies of three proposed Pseudo-Stereo variants and $L_{depth}$ at IOU threshold 0.7. Exp. is the experiment tag.}
  \vspace{-4mm}
    \label{tab:ablation}
\end{table}

\begin{table*}[t!]
  \small
  \centering
  \resizebox{0.72\textwidth}{!}{
  \begin{tabular}{l|c|ccc|ccc}
    \toprule
    \multirow{2}*{Methods} &\multirow{2}*{Reference}  &\multicolumn{3}{c|}{$AP_{3D}$} & \multicolumn{3}{c}{$AP_{BEV}$}\\
    {}  & {} & Easy & Moderate & Hard & Easy & Moderate & Hard \\
    \midrule
    MonoDIS\cite{simonelli2019disentangling}  &ICCV 2019  &10.37 &7.94 &6.40 &17.23 &13.19  &11.12 \\
    AM3D\cite{ma2019accurate}  &ICCV 2019    &16.50 &10.74 &9.52 &25.03 &17.32  &14.91 \\
    M3D-RPN\cite{brazil2019m3d}      &ICCV 2019   &14.76 &9.71 &7.42 &21.02 &13.67 &10.23  \\
    D4LCN\cite{ding2019learning}      &CVPR 2020  &16.65 &11.72 &9.51 &22.51  &16.02  &12.55 \\
    MonoPair\cite{chen2020monopair}  &CVPR 2020  &13.04 &9.99 & 8.65 &19.28 &14.83 &12.89 \\
    MonoFlex\cite{zhang2021objects} &CVPR 2021   &19.94 &{13.89} &{12.07} &28.23   &{19.75}  &16.89 \\
    MonoEF\cite{zhou2021monocular}     &CVPR 2021     &{21.29} &13.87 &11.71 &{29.03} &19.70 &{17.26} \\
    GrooMeD-NMS\cite{kumar2021groomed}      &CVPR 2021   &18.10 &12.32 &9.65 &26.19 &18.27 &14.05 \\
    CaDDN\cite{reading2021categorical}     &CVPR 2021 &19.17 &13.41 &11.46 &27.94 &18.91 &17.19 \\
    DDMP-3D\cite{wang2021depth}      &CVPR 2021     &19.71 &12.78 &9.80 &28.08   &17.89  &13.44 \\
    MonoRUn\cite{chen2021monorun}     &CVPR 2021   &19.65 &12.30 &10.58 &27.94 &17.34 &15.24 \\
    DFR-Net\cite{zou2021devil}     &ICCV 2021    &19.40 &13.63 &10.35 &28.17 &19.17 &14.84 \\
    MonoRCNN\cite{shi2021geometry}     &ICCV 2021    &18.36 &12.65 &10.03 &25.48 &18.11 &14.10 \\
    DD3D\cite{park2021pseudo} &ICCV 2021   &{23.22} &{16.34} &{14.20} &{30.98} &{22.56} &{20.03} \\
    \midrule
    Ours-im   &--     &19.79 &13.81 &12.31 &28.37  &20.01 &17.39 \\
    Ours-fld  &--    &\textbf{23.74} &\textbf{17.74} &\underline{15.14} &\textbf{32.84}  &\textbf{23.67}  &\textbf{20.64} \\
    Ours-fcd &--   &\underline{23.61} &\underline{17.03} &\textbf{15.16} &\underline{31.83}  &\underline{23.39}  &\underline{20.57} \\
    \bottomrule
  \end{tabular}
  }
  \vspace{-2mm}
  \caption{Performance for \emph{Car} of three methods on KITTI \emph{test} at IOU threshold 0.7. The best results are \textbf{bold}, the second best \underline{underlined}.}
  \label{tab:kitti_test_server}
  \vspace{-3mm}
\end{table*}

\subsection{Ablation Study}
As shown in Table.~\ref{tab:ablation}, we conduct ablation studies on the KITTI \emph{val} for the three proposed Pseudo-Stereo variants: image-level, feature-level and feature-clone generation. 
%We investigate the effects of the estimated depth map and the depth loss.

\noindent\textbf{Image-level.} As shown in Exp.1 and Exp.2 in Table.~\ref{tab:ablation}, with the depth loss, the overall performance of the image-level generation method decreases by (-0.38\%, -0.83\%, -0.86\%) for $AP_{3D}$ and (-1.05\%, -1.16\%, -1.17\%) for $AP_{BEV}$. The pixel-correspondence constraints of the generated Pseudo-Stereo pairs may have large offsets against the ground truth because of the depth estimation errors. Learning with the virtual right image warped from the inaccurate pixel-correspondences causes the feature degradation. Forcing the degraded feature to 
fit the ground-truth depth map with the depth loss impairs the overall performance. %because of the huge gap between the ground-truth depth map and the degraded feature.

\noindent\textbf{Feature-level.} As can be seen from Exp.3 and Exp.4 in Table.~\ref{tab:ablation}, the feature-level generation with the depth loss achieves significant improvements on $AP_{3D}$ (\textbf{+13.04\%}, \textbf{+7.97\%}, \textbf{+6.04\%}) and $AP_{BEV}$ (\textbf{+14.4\%}, \textbf{+9.48\%}, \textbf{+7.01\%}). The forward-warping used in image-level generation is a non-learning process from image to image, while the feature-level generation is an adaptive and differentiable learning process with disparity-wise dynamic convolutions from feature to feature. The virtual right features are generated from the left features and the disparity features. Thus, the feature degradation caused by the depth estimation errors is mitigated by embedding the estimated depth information into high dimensional feature space and using the embedded depth representations to filter the left features dynamically in the feature-level generation. The gap between the ground-truth depth and the feature is mitigated. With the extra depth guidance from the depth loss, the depth representations can strengthen the embedded 3D measurements in feature-level, achieving much better performance.
%\subsection{Qualitative results}

\noindent\textbf{Feature-clone.} From Exp.5 and Exp.6 in Table.~\ref{tab:ablation}, it shows that the feature-clone achieves significant improvements with the depth loss on $AP_{3D}$ (+4.13\%, +2.06\%, +1.95\%) and $AP_{BEV}$ (+4.67\%, +2.01\%, +1.66\%). This lies in the fact that feature-clone does not require depth estimation network and the depth loss alone can improve the awareness of depth information in features. 

%By comparing the performance feature-level and feature-clone methods, we find that the depth loss is essential in feature-level generation for our framework..

\begin{table*}[t!]
    \centering
    \resizebox{1.7\columnwidth}{!}{
    \begin{tabular}{l|ccc|ccc}
    \toprule
    \multirow{2}*{Methods}  & \multicolumn{3}{c|}{$Pedestrian \quad AP_{3D}/AP_{BEV}$}  & \multicolumn{3}{c}{$Cyclist \quad AP_{3D}/AP_{BEV}$}\\
    {}                    &Easy   &Moderate &Hard    &Easy  &Moderate  &Hard\\
    \midrule
    D4LCN\cite{ding2019learning}             &4.55 / 5.06   &3.42 / 3.86   &2.83 / 3.59   &2.45 / 2.72   &{1.67} / 1.82   &1.36 / {1.79}\\
    MonoPSR\cite{ku2019monocular}      &6.12 / 7.24 &4.00 / 4.56 &3.30 / 4.11&8.37 / 9.87 &4.74 / 5.78 &3.68 / 4.57\\
    CaDDN\cite{reading2021categorical}      &12.87 / 14.72 &8.14 / 9.41 &6.76 / 8.17 &7.00 / 9.67 &3.41 / 5.38 &3.30 / 4.75\\
    MonoFlex\cite{MonoFlex}      &9.43 / 10.36 &6.31 / 7.36 &5.26 / 6.29 &4.17 / 4.41 &2.35 / 2.67 &2.04 / 2.50\\
    GUPNet\cite{lu2021geometry}      &\underline{14.95} / {15.62} &\underline{9.76} / {10.37} &\underline{8.41} / {8.79} &5.58 / 6.94 &3.21 / 3.85 &2.66 / 3.64\\
    \midrule
    Ours-im &8.26 / 9.94  &5.24 / 6.53   &4.51 / 5.72   &4.72 / 5.76   &2.58 / 3.32   &2.37 / 2.85\\
    Ours-fld &\textbf{16.95} / \textbf{19.03} &\textbf{10.82} / \textbf{12.23} &\textbf{9.26} / \textbf{10.53}&\textbf{11.22} / \textbf{12.80}&\textbf{6.18} / \textbf{7.29}&\textbf{5.21} / \textbf{6.05}\\
    Ours-fcd & 14.33 / \underline{17.08}  &9.18 / \underline{11.04}   &7.86 / \underline{9.59}   &\underline{9.80} / \underline{11.92}   &\underline{5.43} / \underline{6.65}   &\underline{4.91} / \underline{5.86}\\
    \bottomrule
    \end{tabular} }
    \vspace{-3mm}
    \caption{Performance for \emph{Pedestrian} and \emph{Cyclist} on KITTI \emph{test} at IOU threshold 0.5. The best results are \textbf{bold}, the second best \underline{underlined}.}
    \vspace{-1mm}
    \label{tab:ped}
\end{table*}

\begin{table}[t!]
    \small
    \centering
    \resizebox{0.64\columnwidth}{!}{
    \begin{tabular}{l|ccc}
    \toprule
    \multirow{2}*{Methods}  & \multicolumn{3}{c}{$AP_{3D}$}  \\
    {} &Easy   &Moderate &Hard \\
    \midrule
    D4LCN\cite{ding2019learning} & {22.32} & {16.20} & {12.30} \\
    DDMP-3D\cite{wang2021depth} & {28.12} & {20.39} & {16.34} \\
    CaDDN\cite{reading2021categorical} & {23.57} & {16.31} & {13.84} \\
    MonoFlex\cite{zhang2021objects} & {23.64} & {17.51} & {14.83} \\
    GUPNet\cite{lu2021geometry}& {22.76} & {16.46} & {13.72} \\
    \midrule
    Ours-im&\underline{31.81} & \underline{22.36} & \underline{19.33} \\
    Ours-fld& \textbf{35.18} & \textbf{24.15} &\textbf{20.35} \\
    Ours-fcd& {28.46} & {19.15} & {16.56} \\
    \bottomrule
    \end{tabular} }
    \vspace{-3mm}
  \caption{Performance for \emph{Car} on KITTI \emph{val} set at IOU threshold 0.7. The best results are \textbf{bold}, the second best \underline{underlined}.}
    \label{tab:kitti_val}
    \vspace{-4mm}
\end{table}

%From the comparison between image-level and feature-level generation methods in Table.~\ref{tab:ablation}, the adaptive feature conversion for virtual view and embedding the depth information into high dimensional feature space with disparity-wise dynamic convolution can avoid the feature degradation, which is caused by forcing the network to learn with the virtual right image warped from the inaccurate pixel-correspondences. 
\noindent\textbf{Estimated Depth Map.} From the comparison among Exp.2, Exp.4 and Exp.6 in Table.~\ref{tab:ablation}, with the estimated depth map used in both image-level (Exp.2) and feature-level (Exp.4), the models achieve better performance than the model without using the estimated depth map (Exp.6), which implies that the estimated depth map is effective in both image-level and feature-level in our framework. In image-level, the degraded feature caused by inaccurate pixel-correspondences is not forced to fit the ground-truth depth without the depth loss, and the estimated depth map improves the capability of perceiving depth information in the image input level, improving performance for monocular 3D detection. In feature-level, the feature degradation is eased by the proposed DDC in high dimensional feature space and the estimated depth map improves the capability of perceiving depth information in features, thereby achieving better performance than the image-level methods.

\noindent\textbf{DDC.} By comparing the performance feature-level and feature-clone methods with and without depth loss, we find that the depth loss is essential in feature-level generation to monocular 3D detection. We use the feature-clone method with depth loss as the baseline and add DDC to the baseline (Exp.3 in Table.~\ref{tab:ablation}) to shown the effect. Feature-level generation with the proposed DDC achieves significant improvements against the baseline, indicating that the proposed DDC is effective in generating the virtual right view in feature level for monocular 3D detection. This lies in the fact that the proposed DDC uses the embedded depth representations to dynamically filter the left features, deriving depth-aware feature learning and achieving significant improvements in monocular 3D detection.

%the strategy II with depth loss outperforms the strategy I without depth loss with margins on $AP_{3D}$ (+3.37\%, +1.79\%, +1.02\%) and $AP_{BEV}$ (+2.63\%, +1.87\%, +1.19\%). In the one hand, the influence of the wrong pixel-correspondence constraint caused by the error of depth map cannot be reduced by adding a depth loss. In the other hand, the process of generating the virtual right image is separated from the network so cannot be optimized. Another problem that cannot be ignored is that the process of generating the virtual right image is a time-consuming process. Compared to generating the virtual right image, generating the virtual right feature doesn't need an additional process for generating the virtual right image. It's also possible to reduce the impact of depth map errors through an additional depth loss.

\begin{figure*}[t!]
  \centering
 \includegraphics[width=1\textwidth]{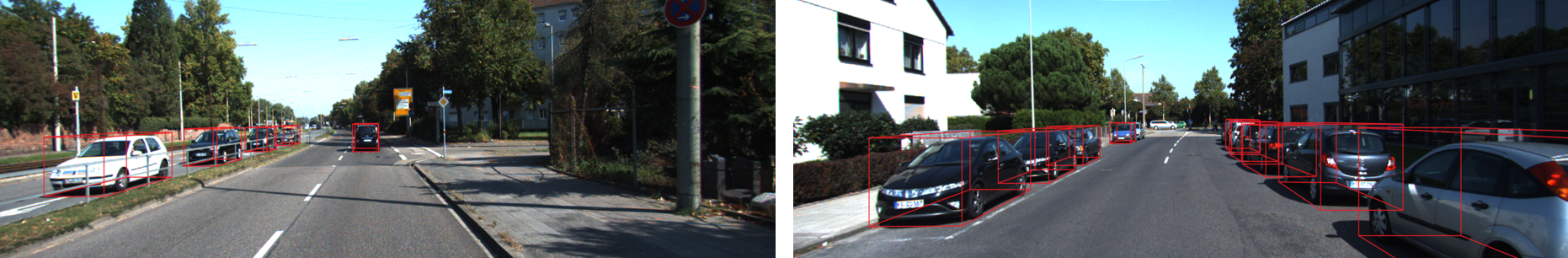}
 \vspace{-4mm}
  \caption{Qualitative results of the best model (Ours-fld) on KITTI \emph{val} set with red 3D bounding boxes.} 
 \label{fig:result}
 \vspace{-2mm}
\end{figure*} 

\subsection{Quantitative and Qualitative Results}
We evaluate the three proposed Pseudo-Stereo variants: image-level generation, feature-level generation and feature clone, on KITTI \emph{test} and \emph{val} set. 
From the above ablation studies, we choose the strategy with better performance for each method: image-level generation without depth loss (Ours-im), feature-level generation with depth loss (Ours-fld) and feature-clone with depth loss (Ours-fcd).

%\textbf{Results on \emph{test} set.}
%\textbf{Quantitative results on KITTI \emph{test} and \emph{val} set.}
\noindent\textbf{Results on \emph{test} set.} Table.~\ref{tab:kitti_test_server} shows the performance comparison for \emph{Car} on KITTI \emph{test} server and Table.~\ref{tab:ped} shows the performance comparison for \emph{Pedestrian} and \emph{Cyclist} on KITTI \emph{test} server. DD3D \cite{park2021pseudo}, GUPNet \cite{lu2021geometry} and MonoPSR \cite{ku2019monocular} rank 1$^{st}$ on \emph{Car}, \emph{Pedestrian} and \emph{Cyclist}, respectively, for monocular 3D detection in KITTI-3D benchmark before this work. As shown in Table.~\ref{tab:kitti_test_server} and Table.~\ref{tab:ped}, Ours-fld achieves better performance than DD3D \cite{park2021pseudo}, GUPNet \cite{lu2021geometry} and MonoPSR \cite{ku2019monocular} across all three object classes on both $AP_{3D}$ and $AP_{BEV}$ for monocular 3D detection using single model only. Moreover, our three methods achieve 18/18 best results and 15/18 second-best results across all three object classes on both $AP_{3D}$ and $AP_{BEV}$. Note that Ours-fld achieves 17 out of 18 best results except the hard level of car, where the best is Our-fcd. This implies that the proposed Pseudo-Stereo 3D detection framework is very effective in monocular 3D detection. 

\noindent\textbf{Results on \emph{val} set.} As shown in Table.~\ref{tab:kitti_val}, Ours-fld outperforms state-of-the-art methods on KITTI \emph{val} set. Figure~\ref{fig:result} shows the qualitative results of Our-fld, the best model, on KITTI \emph{val} set.

\noindent\textbf{Generalization Capability.} Usually, there is a large gap between the monocular 3D detection performance on val set and test set because of over-fitting. As shown in Table.~\ref{tab:kitti_test_server} and Table.~\ref{tab:kitti_val}, the performance gap for Ours-fcd is much smaller than Ours-im and Ours-fld. This lies in the fact that feature-clone method does not require the estimated depth map for training, leading to better generalization capability. Note that we provide both options in our framework.

\section{Conclusion}
We propose a Pseudo-Stereo 3D detection framework with three novel virtual view generation methods, including image-level generation, feature-level generation and feature-clone, for detecting 3D objects from a single image, achieving significant improvements in monocular 3D detection. The proposed framework with our feature-level virtual view generation method ranks 1$^{st}$ among the monocular 3D detectors with publications across three object classes on KITTI-3D benchmark. In feature-level virtual view generation, we propose a disparity-wise dynamic convolution with dynamic kernels from disparity feature map to filter the features adaptively from a single image for generating virtual image features, which eases the feature degradation and achieves significant improvements.
We analyze two major effects of depth-aware feature learning in our framework. 

\noindent\textbf{Broader Impacts.} Our Pseudo-Stereo 3D detection framework has the potential to provide a new perspective of monocular 3D detection with Pseudo-Stereo views to our community. Also, our analysis of depth-aware feature learning in Pseudo-Stereo frameworks may give an inspiration to mitigate the performance gap between monocular and stereo 3D detectors.   

\textbf{Acknowledgement:} This work was supported by
the National Key Research and Development Program of
China (2018YFE0183900). Hang Dai would like to thank
the support from MBZUAI startup fund (GR006).

%%%%%%%%% REFERENCES
{\small
\bibliographystyle{ieee_fullname}
\bibliography{egbib}
}

\end{document}